\documentclass[10pt,twocolumn,letterpaper]{article}

\usepackage[pagenumbers]{iccv} 

\usepackage{subcaption}
\usepackage{multirow}
\usepackage{floatrow}
\usepackage{caption}
\usepackage{comment}

\usepackage{amsmath}
\usepackage{amssymb}
\usepackage{xcolor}
\usepackage{graphicx}
\usepackage{wrapfig}
\usepackage{array}
\usepackage{makecell}
\usepackage{color}
\usepackage{colortbl}
\usepackage{booktabs} 

\definecolor{iccvblue}{rgb}{0.21,0.49,0.74}
\usepackage[pagebackref,breaklinks,colorlinks,allcolors=iccvblue]{hyperref}

\def\paperID{*****} 
\def\confName{ICCV}
\def\confYear{2025}

\title{Generalizable Prompt Learning of CLIP: A Brief Overview}

\author{
Fangming Cui\\
{\tt\small cuifangming@sjtu.edu.cn}
\and
Liang Xiao\\
{\tt\small xiaoliang.cs@gmail.com}
\and
Xule Wang\\
{\tt\small wangxule@meituan.com}
\and
Xuan Wang\\
{\tt\small wangxuan39@meituan.com}
\and
Yonggang Zhang\\
{\tt\small ygzhang@comp.hkbu.edu.hk}
}

\begin{document}
\maketitle

\begin{abstract}
Existing vision-language models (VLMs) such as CLIP have showcased an impressive capability to generalize well across various downstream tasks. These models leverage the synergy between visual and textual information, enabling them to understand and reason about the content present in images and text in a unified manner. This article provides a brief overview of CLIP based on few-shot prompt learning, including experimental data and technical characteristics of some methods. The purpose of this review is to provide a reference for researchers who have just started their research in generalizable prompting of CLIP through few-shot training for classification across 15 datasets and also to facilitate the integration of this field by researchers in other downstream tasks. 
\end{abstract}

\section{Introduction}
\label{sec:intro}

Large-scale pre-trained models, such as BERT~\cite{bert} and T5~\cite{t5}, have made significant progress in the fields of natural language processing and computer vision. Large-scale pre-training models in the field of computer vision, such as ViT~\cite {vit}, treat images as sequential data and learn a large amount of general visual knowledge through pre-training on large-scale image datasets. These advancements have laid the foundation for the emergence of multimodal pre-training models for image segmentation~\cite{luo2023segclip,luo2018coarse,cao2024towards,ding2023maskclip,ding2023open}, 3D detection~\cite{luo2023kecor,greer2024and,chen2023towards,etchegaray2024find}, visual question answering~\cite{yu2018beyond,yu2023anetqa,yu2023bilaterally,yu2019deep,yu2019activitynet,yu2017multi}, and other applications~\cite{liu2024motion,yang2022diversity,shi2024foodfusion,liu2024decoupling,wang2021spgnet,liu2023fsi,li2021learning,liuzhe8}.
\begin{table*}[h]
\footnotesize
		\caption{An overview of designs. Base-to-novel generalization tasks.}		
		\centering
		\begin{tabular}{c|l|c|c|c|c|c|c} 
			\toprule 
			 &Method&Publication&Prompting &Contribution& Base& Novel&HM \\
			\midrule 
      1&CLIP~\cite{radford2021learning} &(ICML2021)& No learning&Zero-shot and strong generalization. capability & 69.34&\textbf{74.22}&71.70 \\
            \midrule 
       \rowcolor{gray!20} 2 &CoOp~\cite{zhou2022learning} &(IJCV2022) &T& First proposed prompt learning method. & 82.63 &67.99& 74.60    \\
	 \rowcolor{gray!20}3&CoCoOp~\cite{zhou2022conditional}& (CVPR2022)&T& Focous on generalization problem.& 80.47 &71.69& 75.83   \\
       \rowcolor{gray!20}4  &PLOT~\cite{chen2022plot}& (ICLR2023)&T + V& Applying optimal transport for prompting field.& 81.24 &72.98& 76.89 \\  
       \rowcolor{gray!20}5  &KgCoOp~\cite{yao2023visual}& (CVPR2023)&T&Applying hand-crafted prompts and tuning prompts. &80.73&73.60& 77.00 \\
       \rowcolor{gray!20} 6   &ProGrad~\cite{zhu2023prompt}&(ICCV2023) &T&Appling KL loss for multiple logits.  & 82.48&70.75& 76.16 \\
       \rowcolor{gray!20} 7   &AAPL~\cite{kim2024aapl} & (CVPR2024) & T& Adding attributes to prompt learning.&80.27&72.17& 76.01\\
       \midrule
       8 &MaPLe~\cite{2023maple}& (CVPR2023) &T + V&Multi-modal prompt learning and prompt shifting. &82.28&75.14& 78.55 \\
      9  &PromptSRC~\cite{khattak2023self} & (ICCV2023)&T + V &Three-pronged self-regulating without forgetting.&84.21&75.75& 79.97  \\
       10 &ALIGN~\cite{NEURIPS2023_a547d869} & (NIPS2023)&T + V& Developing token-level optimal transport for prompting. &83.38&75.51& 79.25 \\
       11 &QNet~\cite{shi2023prompt} & (ICLR2024) & T + V& Applying quaternion networks for prompting.&83.32&75.65& 79.30\\
      12  &CoPrompt~\cite{roy2023consistency} & (ICLR2024)&T + V& LLM + Adapter + image augmentations. &84.00&77.23& 80.48\\
       13 &DePT~\cite{zhang2024dept} &(CVPR2024)&T or T + V & A Plug-and-play design.  & 85.19&76.17& 80.43 \\
       14 &TCP~\cite{yao2022prompt} & (CVPR2024) &T&  Incorporating prior
knowledge of each class into embeddings.&84.13&75.36& 79.51\\
       15 &ProMetaR~\cite{park2024prompt} & (CVPR2024) &T + V&Prompt learning via meta-regularization. &84.39&76.93&80.49 \\
       16 &TriMPL~\cite{liu2024trimpl} & (ICMR2024) & T& Masked multi-prompt learning with knowledge mixing.&82.19&74.77& 78.30\\
      17  &CPL~\cite{zhang2025consistent}& (KBS2025)&T + V& Adversarial training prompts + hand-crafted textual prompts. &84.90&77.11 &80.66\\        
		\bottomrule 
		\end{tabular}
  \label{design}
\end{table*}
Researchers have begun exploring how to integrate language models and visual models to achieve a unified understanding of text and images. This has led to the emergence of many multimodal pre-training models for graphics and text, also known as Vision Language Models (VLMs), such as CLIP~\cite{radford2021learning} and ALIGN~\cite{ALIGN}. These models are trained on large-scale image text pairs using strategies such as contrastive learning and masked language modeling to learn cross-modal representations of text and images. For example, CLIP trains 400 million pairs of image texts to jointly learn the representations of images and texts by maximizing the similarity of identical pairs and minimizing the similarity between non-paired image texts. These VLMs demonstrate powerful joint representation capabilities, but migrating them to downstream tasks is a new challenge~\cite{luo2020adversarial}. Fine-tuning usually requires a large amount of labeled data to adjust model parameters, which is not suitable for migration in small sample scenarios. Meanwhile, fine-tuning may disrupt the general knowledge learned by VLM during the pre-training phase, leading to catastrophic forgetting. The pioneering research in the field of few-shot image recognition can be traced back to the era of manual features. With the rise of deep learning~\cite{bengio2021deep}, researchers have begun to focus on how to use the powerful fitting and generalization abilities of neural networks to solve few-shot image recognition problems~\cite{piratla2020efficient,li2018deep,muandet2013domain,rame2023model}. 

In recent years, as CLIP has opened up a new paradigm for image classification, efficiently transferring CLIP to image classification tasks in zero-shot and few-shot scenarios has become a research hotspot. 
The CLIP has demonstrated excellent classification performance in few-shot image classification tasks, thanks to its use of hand-crafted prompt engineering techniques to construct text prompts for each classification category, and then extract feature vectors using text and image encoders to calculate cosine similarity for image classification. These prompts can include hand-engineered text, such as ``a photo of a [class]'', which guides the text encoder of CLIP. Compared to retraining the model, CLIP's prompt engineering method allows for the freedom to change the number of categories, creating a new paradigm for image classification. 
After CLIP, some studies~\cite{zhou2022conditional,lu2022prompt} have been inspired by prompt learning in the NLP field, using learnable prompts instead of fixed templates "a photo of a [CLASS]" for training on small sample data, achieving better classification results for designated samples. The introduction of learnable prompts in the image modality of CLIP is to adjust the feature representation of the image to be closer to the text prompt feature vector of the corresponding category, thereby improving classification accuracy. For a pioneering perspective, CoOp~\cite{zhou2022learning} introduced learnable prompts for the first time in task transfer in CLIP, replacing the "a photo of a [CLASS]" template with learnable text prompts, which significantly improved image classification tasks compared to CLIP. 
However, CoOp performs poorly in generalization, and the learned prompts are ineffective in novel classes (Table~\ref{design}), it could be the overfitting of training images~\cite{zhou2022conditional}.  These methods often rely on a learnable prompt and cannot fully describe the classification categories, thereby limiting their ability to distinguish between different distributions~\cite{li2019feature,cha2021swad,carlucci2019domain,volpi2018generalizing}.
The research on extending prompt learning to the imaging modality of CLIP has not fully considered the synergistic effect between text prompt learning and images and has not fully utilized the rich multimodal~\cite{yu2020deep} knowledge contained in CLIP~\cite{zhou2022conditional}.
To address this issue, CoCoOp~\cite{zhou2022conditional} combines image feature vectors with learnable text prompts based on CoOp, enhancing attention to image instance information and improving classification accuracy on unseen categories. 
Many studies have been devoted to effectively applying prompt learning for CLIP in few-shot image classification tasks, but these works still have main issues.
In table~\ref{design}, CoCoOp's performance in Novel classes surpasses that of CoOp's, yet it still falls short of the CLIP with no learning prompts, and these models~\cite{chen2022plot,yao2023visual,zhu2023prompt,kim2024aapl,lu2022prompt} also have similar issues.

\begin{figure*}[ht]
\centerline{\includegraphics[scale=0.37]{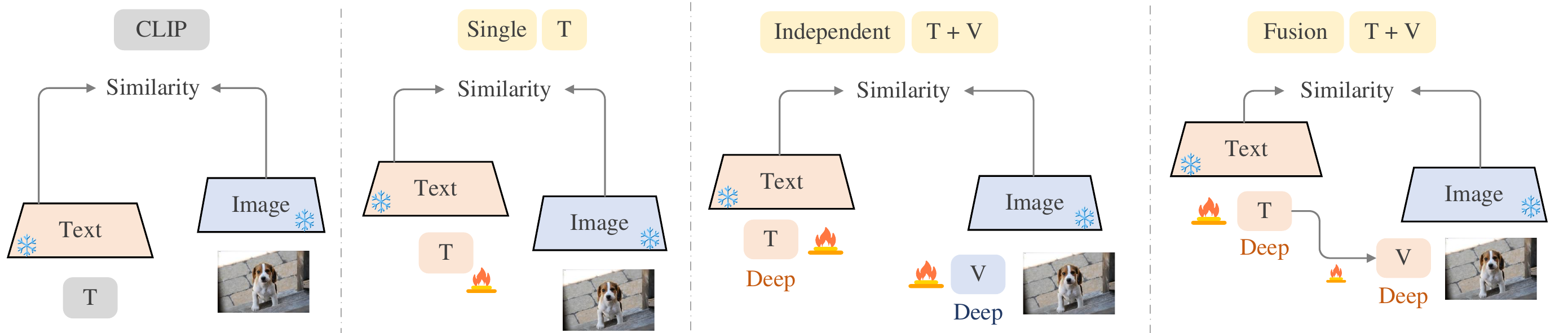}}
\caption{An overview of representative framework designs.}    
\label{compare}
\end{figure*}

\section{Generalizable Prompt Learning}
In the early stages of the development of pre-trained models, fine-tuning is a commonly used method for applying pre-trained visual language models~\cite{huang2022unsupervised} to downstream tasks. The base idea of fine-tuning is to add lightweight network structures specific to downstream tasks to the pre trained model, and then design appropriate training tasks on labeled data specific to downstream tasks to adjust the parameters of the added network structure or the entire pre-trained model. Although fine-tuning has performed well in many tasks, there are also some limitations.
Fine tuning requires a large amount of downstream task specific labeled data, which is often difficult to obtain.
Designing suitable network structures and training tasks for each downstream task requires a significant amount of computing resources and time.
The mismatch between fine-tuning tasks and pre training tasks may lead to catastrophic forgetting of pre trained models.
These limitations make fine-tuning methods less effective or feasible in certain situations~\cite{liu2024trimpl}.

Therefore, researchers have begun to explore new transfer learning methods to overcome the limitations of fine-tuning methods, such as zero-shot learning~\cite{radford2021learning,goyal2023finetune,shu2022test}, transfer learning~\cite{pham2023combined}, meta learning~\cite{li2018learning,park2024prompt}, etc. These methods aim to achieve more efficient and flexible transfer learning and generalization tasks~\cite{dai2023moderately,shu2023clipood,chen2022learning,pengconjnorm,jiang2024negative,chen2023understanding,kumar2022fine,chattopadhyay2020learning}, in order to successfully apply pre-trained models to various downstream tasks without the need for large amounts of labeled data and tedious task specific fine-tuning.

On this basis, researchers began to study the use of continuous textual prompts to replace manually constructed discrete prompts~\cite{li2021prefix, nie2023out, cao2023domain, nie2023out} in the textual branch. Continuous prompts, also known as learnable prompts, consist of a sequence of learnable vectors. The opposite of discrete prompts is continuous prompts, which do not rely on manually designed templates or prompt words, but instead train and optimize learnable vectors on small sample data from downstream tasks. This enables the model to automatically generate the most suitable prompts for the current task during the learning process, thereby improving the adaptability and performance of the model~\cite{liu2021p, lester2021power}.
For example, prefix tuning adds task-specific learnable vector sequences before input, optimizing the learnable vector sequences on task-specific few sample data to avoid manual design prompts. This method has achieved better results in multiple downstream tasks~\cite{cui1,cui2,cui3}, demonstrating the potential of continuous prompting in improving model adaptability and performance.

The prompt-based methods mentioned above mainly target the text modality of CLIP. However, as a multimodal model of text and image, CLIP's image modality is also crucial in downstream tasks. Researchers have attempted to successfully apply prompt learning methods to the field of image encoder, known as VPT (Visual Prompt Tuning)~\cite{jia2022visual}. The overall architecture of VPT draws on the idea of prefix tuning and applies learnable prompts to image block sequences to transfer Vision Transformer (ViT) to downstream tasks through prompt learning. VPT-deep adds learnable embeddings to each Transformer layer of ViT. This method enables the model to better adapt to various computer vision tasks through prompt learning, demonstrating the potential and effectiveness of prompt learning in the field of computer vision~\cite{2023maple,khattak2023self,chen2022plot,cao2023domain,cao2024aaai,icml2024}.
Recently, the powerful generalization ability of CLIP~\cite{radford2021learning} has made it the foundation for many methods that apply pre-trained VLMs to downstream tasks. Prompt Learning~\cite{liu2021p, li2021prefix} is a widely used technique in NLP for learning downstream tasks. It involves merging task-specific information into the input token. Using text prompts, which provide instructions to the Visual-Language Model (VLM) language branch, is a common practice for enhancing task understanding. However, the limitations of these methods have stimulated research on new technologies inspired by rapid tuning in natural language processing (NLP).

CoOp~\cite{zhou2022learning} fine-tunings the CLIP model specifically for low sample image recognition by optimizing a set of continuous label embeddings in the language branch. Inspired by the use of learnable prompts instead of discrete prompts in NLP and CV fields, CoOp replaces the manually constructed prompts in CLIP with learnable prompts. CoOp uses a small amount of image classification data to optimize the representation of learnable prompts, achieving excellent classification performance on low-sample image classification tasks. In the overall architecture of CoOp, the text encoder and image encoder remain frozen during the training process, and only learnable prompts are optimized. Research has found that trained prompts do not have language meanings that are understandable to humans when mapped to text sequences, indicating the necessity of using learnable prompts instead of manually constructed prompts. The learnable prompts are trained through supervised data, breaking the limitations of human language knowledge on prompt design.

The image conditional prompts of CoCoOp~\cite{zhou2022conditional}  significantly enhance the generalization ability to unknown classes and reduce the risk of overfitting to limited labeled data.  CoCoOp proposes a conditional prompt learning method that can adaptively generate prompts based on the current input sample, thereby achieving better cross-distribution generalization performance. By conditioning the prompts into visual features, CoCoOp ensures that the language model pays attention to relevant visual information when generating predictions. This conditioned reflex helps improve the model's generalization ability to unknown examples and reduces the risk of overfitting to limited labeled data. 

CoPL~\cite{copl} points out that introducing all the information of the image into the prompt may introduce a large amount of noise. Therefore, it enhances the weight of image information related to the prompt by calculating the similarity between local features of the image and learnable prompts and weighing them. 

KgCoOp~\cite{yao2023visual} and PromptSRC~\cite{khattak2023self} found that while optimizing learnable prompts using few sample data, reducing the distance between learnable prompts and "a photo of a [CLASS]" helps balance the fitting and generalization abilities of the prompts. ProGrad~\cite{zhu2023prompt} constrains the gradient update direction of learnable prompts through zero sample prediction results, taking into account both knowledge in few samples and generalization knowledge in ``a photo of a [CLASS]" to alleviate overfitting problems. These methods rely on a single prompt and cannot provide detailed and comprehensive matching for complex image data. To address this issue, ProDA~\cite{lu2022prompt} constructs diverse descriptions for each category using distribution learnable prompts to match rich and diverse image information, but ProDA does not consider the generalization performance of multiple prompts.

MaPLe~\cite{2023maple} introduces learnable text prompts in the text modality of CLIP and converts them into image prompts through linear layers to better align text and visual features. However, these methods did not fully utilize the symmetry between image and text modalities in CLIP, nor did they consider the synergistic effect between text prompt learning and image prompt learning.  
To tackle this issue, QNet~\cite{shi2023prompt} employs the quaternion networks in multi-modal prompting for generalization tasks. QNet utilizes a quaternion hidden space in which the mutually orthogonal imaginary axes capture diverse intermodal semantic spatial correlations from multiple perspectives.
In the literature inspired by prompt learning~\cite{zhu2023prompt}, it is recommended to learn a single text prompt to guide the visual language pre-training model, instead of manually designed prompts. In addition, some methods~\cite{yao2023visual, lu2022prompt, Yao_2023_CVPR,li2025divergenceenhanced}
constrain the proposed learnable prompts to include base common sense and prior distribution learning.  

In addition to single modal prompt tuning, some methods have also introduced multimodal prompt tuning design in CLIP to effectively align V-L representations. Method~\cite{chen2022plot} adopts the optimal transport~\cite{villani2009optimal,redko2019optimal,cuturi2013Sinkhorn} technology for multimodal fusion and image classification tasks. Further, ALIGN~\cite{NEURIPS2023_a547d869} employs the optimal transport for token-level prompting. However, their method has many parameters and has not fully unleashed the potential generalization ability of CLIP. In addition, there are several methods~\cite{yao2023visual,lu2022prompt,zhu2023prompt,Yao_2023_CVPR,lu2023beyond,xia2023hgclip} including using learnable prompts that incorporate base knowledge and prior distribution learning. Some methods use adversarial training techniques~\cite{madry2017towards,zhang2024robust,zhang2020principal,pmlr-v119-zhang20o,zhang2021causaladv} for language models or visual models, as well as for fine-tuning prompt word learning in the field of multimodal visual language models~\cite{yang2024revisiting,roy2023consistency,yang2024prompt,das2024human,zhang2024promptfix,zhou2024few,li2024one,zhang2025consistent}.

Recently, a plug-and-play method DePT~\cite{zhang2024dept} has been proposed. DePT is suitable for various modal prompt learning architectures (CoOp, CoCoOp, MaPLe, PromptSRC). The improvement of DePT on CoOp hardly increases the training time, but there is a certain degree of increase in the number of parameters. In Table~\ref{design}, the context learned through traditional image augmentation is biased toward seen classes, negatively impacting generalization to unseen classes~\cite{kim2024aapl}. 
To tackle this issue, AAPL~\cite{kim2024aapl} suggests employing adversarial token embedding to separate low-level visual augmentation features from high-level class information when introducing bias into learnable prompts.

\section{Few-shot Classification}
In the task of few-shot image classifications, there is only a very limited amount of image data available for the model to learn for each category. This setting simulates the process of human learning, where new categories can be understood and recognized by observing a small number of examples~\cite{chen2022federated}. This type of task is very common in the real world, such as in medical imaging and military high-tech tasks, where it may not be possible to obtain a large amount of labeled image data. To address the challenges posed by small sample sizes, researchers have proposed various methods~\cite{balaji2018metareg,zhang2022rich,rame2022diverse,arpit2022ensemble,krueger2021out}.
One method is transfer learning, which transfers the knowledge of a pre-trained model from one task to another, thereby utilizing existing knowledge to enhance its adaptability to new tasks. CLIP uses manually constructed prompts to solve zero sample image classification tasks, which is a transfer learning method.
\begin{table*}[h]
\centering
\footnotesize
\begin{tabular}{lcccccccccccc}
\toprule
          & \textbf{Source} & \multicolumn{11}{c}{\textbf{Target}}          
          \\ \cmidrule(r){2-2} \cmidrule(r){3-13}
          & \multicolumn{1}{c}{\raisebox{-2.5ex}{\rotatebox[origin=c]{90}{\centering ImageNet} }}     
          & \multicolumn{1}{c}{\raisebox{-2.5ex}{\rotatebox[origin=c]{90}{Caltech101}}}    
          &\multicolumn{1}{c}{\raisebox{-2.5ex}{\rotatebox[origin=c]{90}{OxfordPets}}}     
          & \multicolumn{1}{c}{\raisebox{-2.5ex}{\rotatebox[origin=c]{90}{StanfordCars}}}  
          &\multicolumn{1}{c}{\raisebox{-2.5ex}{\rotatebox[origin=c]{90}{Flowers102}}}     
          & \multicolumn{1}{c}{\raisebox{-2.5ex}{\rotatebox[origin=c]{90}{Food101}}}       
          & \multicolumn{1}{c}{\raisebox{-2.5ex}{\rotatebox[origin=c]{90}{Aircraft}}}       
          &\multicolumn{1}{c}{\raisebox{-2.5ex}{\rotatebox[origin=c]{90}{SUN397} }}        
          & \multicolumn{1}{c}{\raisebox{-2.5ex}{\rotatebox[origin=c]{90}{DTD} }}          
          & \multicolumn{1}{c}{\raisebox{-2.5ex}{\rotatebox[origin=c]{90}{EuroSAT}}}       
          & \multicolumn{1}{c}{\raisebox{-2.5ex}{\rotatebox[origin=c]{90}{UCF101} }}       
          & \multicolumn{1}{c}{\raisebox{-2.5ex}{\rotatebox[origin=c]{90}{\textit{Average}} }} \\ 
          
          \midrule

CLIP~\cite{radford2021learning}      & 66.72  & 92.94          & 89.07          & 65.29          & 71.30          & 86.11          & 24.87          & 62.62          & 44.56          & 47.69          & 66.77          & 65.12            \\
\midrule

CoOp~\cite{zhou2022learning}      & \textbf{71.51}  & 93.70          & 89.14          & 64.51          & 68.71          & 85.30          & 18.47          & 64.15          & 41.92          & 46.39          & 66.55          & 63.88            \\

CoCoOp~\cite{zhou2022conditional}   & 71.02           & 94.43          & 90.14          & 65.32 & 71.88          & 86.06          & 22.94          & 67.36          & 45.73& 45.37          & 68.21          & 65.74            \\ 

PLOT~\cite{chen2022plot} & 70.15           & 94.60         & 90.23          & 65.41 & 71.97          & 86.32          & 22.87          & 67.22          &44.99 & 46.57         & 68.32          & 65.85
\\

KgCoOp~\cite{yao2023visual} & 70.66           & 93.92         & 89.83          & 65.41 & 70.01          & 86.36          & 22.51          & 66.16          &46.35 & 46.04         & 68.50          & 65.51
\\

MaPLe~\cite{2023maple}   & 70.72           & 93.53         & 90.49          & 65.57 & 72.23          & 86.20          & 24.74          & 67.01          & 46.49 & 48.06          & 68.69          & 66.30            \\

ProGrad~\cite{zhu2023prompt} & 72.24           & 91.52         & 89.64          & 62.39 & 67.87          & 85.40          & 20.16          & 62.47          &39.42 & 43.46         & 64.29       & 62.71   \\

PromptSRC~\cite{khattak2023self}   & 71.27           & 93.60         & 90.25          & 65.70 & 70.25          & 86.15          & 23.90          & 67.10          & 46.87 & 45.50          & 68.75          & 65.81           \\

ALIGN~\cite{NEURIPS2023_a547d869} & 72.03           & 93.91         & 90.55          & 65.84 & 73.75          & 86.40          & 24.95          & 67.59          &46.75 & 47.25       & 69.60          & 66.66   \\

QNet~\cite{shi2023prompt} & 71.10           & 94.30         & 90.80          & 66.00 & 73.10          & 86.10          & 24.20          & 67.90          &46.90 & 51.80         & 68.70          & 66.98
\\

CoPrompt~\cite{roy2023consistency}   & 70.80           & 94.50         & 90.73          & 65.67 & 72.30          & 86.43          & 24.00          & 67.57          &47.07 & 51.90          & 69.73          & 67.00   \\

DePT~\cite{zhang2024dept} & 71.60           & 93.80         & 90.13          & 66.00 & 70.93          & 86.27          & 24.30          & 67.23          &46.60 & 45.83         & 69.10        & 66.02   \\

TCP~\cite{yao2023tcp}   & 71.40           & 93.97        & 91.25          & 64.69 & 71.21          & 86.69          & 23.45          & 67.15          & 44.35& 51.45          & 68.73          & 66.29            \\

AAPL~\cite{kim2024aapl} & 71.37           & 94.17         & 90.73          & 65.10 & 71.67          & 86.00          & 23.03          & 66.80          &44.80 & 41.83          & 69.30          & 65.34 
\\

ProMetaR~\cite{park2024prompt} & 70.85           & 94.18         & 90.00          & 65.11 & 71.33          & 85.65          & 24.01          & 66.12          &45.71 & 46.88         & 65.12        & 65.41  \\

TriMPL~\cite{liu2024trimpl} & 71.30           & 94.44         & 90.08          & 65.64 & 72.24          & 86.25          & 24.38          & 67.30          &46.19 & 47.29         & 68.89        & 66.27   \\

CPL~\cite{zhang2025consistent} & 71.12           & 93.95         & 91.34          & 65.75 & 71.84          & 86.54          & 25.43          & 67.75          &46.51 & 48.87         & 69.85        & 66.78   \\
\bottomrule
\end{tabular}
\caption{Cross-dataset generalization task. These approaches are
trained on ImageNet and tested on 10 unseen datasets.}
\label{cross}
\end{table*} 

\begin{table*}[h]
\centering
\footnotesize
\begin{tabular}{lcccccc}
\toprule
          & \textbf{Source} & \multicolumn{5}{c}{\textbf{Target}}          
          \\ \cmidrule(r){2-2} \cmidrule(r){3-7}
          
          & \multicolumn{1}{c}{\centering ImageNet}      
          & \multicolumn{1}{c}{\centering -V2}   
          &\multicolumn{1}{c}{\centering -S}   
          & \multicolumn{1}{c}{\centering -A}   
          &\multicolumn{1}{c}{\centering -R}            
          & \multicolumn{1}{c}{\centering Average} 
          \\ 
          \midrule
CLIP~\cite{radford2021learning}      & 66.73  & 60.83          & 46.15
          & 47.77
          & 73.96
          & 57.18
           \\
\midrule

CoOp~\cite{zhou2022learning}     & \textbf{71.51}  & 64.2
          & 47.99
          & 49.71
          & 75.21
          & 59.28
           \\

CoCoOp~\cite{zhou2022conditional}    & 71.02           & 64.07
          & 48.75
          & 50.63 & 76.18
          & 59.91
          \\ 

PLOT~\cite{chen2022plot} & 70.15           & 64.17
          & 49.15
          & 50.83 &76.50
          & 60.16
         \\ 

KgCoOp~\cite{yao2023visual} & 70.66           & 64.10
          & 48.97
          & 50.69 &76.70
          & 60.11
         \\  
          
MaPLe~\cite{2023maple}   & 70.72           & 64.07
          & 49.15
          & 50.9 &76.98
          & 60.27
         \\ 

ProGrad~\cite{zhu2023prompt} & 70.45           & 63.35
          & 48.17
          & 49.45 &75.21
          & 59.05
         \\  
         
PromtSRC~\cite{khattak2023self}   & 71.27           & 64.35
          & 49.55
          & 50.90 &77.80
          & 60.65
         \\ 

ALIGN~\cite{NEURIPS2023_a547d869} & 72.03           & 64.64
          & 49.96
          & 50.94 &76.16
          & 60.43
         \\ 

QNet~\cite{shi2023prompt} & 71.10           & 64.30
          & 49.20
          & 51.30 &77.70
          & 60.65
         \\ 

CoPrompt~\cite{roy2023consistency}   & 70.80           & 64.25
          & 49.43
          & 50.50 &77.51
          & 60.42
         \\ 

DePT~\cite{zhang2024dept}   & 71.60           & 64.51
          & 50.15
          & 51.88 &77.18
          & 60.93
         \\ 

TCP~\cite{yao2023tcp}    &   71.40           & 64.05
          & 49.10
          & 50.55 & 77.8
          & 60.37   
          \\ 
          
AAPL~\cite{kim2024aapl} & 71.37           & 64.20
          & 48.80
          & 50.60 &76.87
          & 60.12
         \\ 

ProMetaR~\cite{park2024prompt} & 71.29           & 64.39
          & 49.55
          & 51.25 &77.89
          & 60.77
         \\ 

TriMPL~\cite{liu2024trimpl} & 71.30           & 64.38
          & 49.48
          & 50.80 &77.25
          & 60.48
         \\

CPL~\cite{zhang2025consistent} & 71.12           & 64.40 & 49.55& 50.95 &78.10 & 60.75 \\

\bottomrule
\end{tabular}
\caption{Domain generalization task. }
\label{Domain}
\end{table*} 
There have been many works that use learnable prompts to transfer CLIP to a few-shot sample image classification tasks. These works either improve the problems of CoOp or extend the form of prompt learning, such as extending a single prompt to multiple prompts or extending text prompts to image prompts.
Few-shot learning has brought new challenges to the traditional field of machine learning. In the current research framework, small sample learning learns generalizable models or knowledge on a large amount of additional data and then transfers it to downstream tasks with limited training data~\cite{0ActionCLIP}. This means that there is often a difference in data distribution between the actual downstream task data and the upstream training data. In traditional machine learning problems, researchers typically assume that the training and testing data are independent and identically distributed. However, small sample learning no longer follows this assumption and requires the model to learn the ability to generalize across distributions from upstream training processes, rather than just considering generalization problems in data with the same distribution.

Unlike zero sample image classification~\cite{yi2022exploring,xian2017zero}, the challenge of small sample learning is that the model must have the ability to generalize across distributions~\cite{zhang2025learning}, that is, it can still effectively learn and generalize when facing downstream tasks with different distributions from the training data. In this case, the model needs to be able to capture changes in data distribution, avoid overfitting specific distributions of training data, and learn commonalities and generality between tasks. Therefore, small sample learning requires models to be more flexible and adaptable, able to quickly adapt to new tasks and data distributions, and thus more effectively apply and generalize in the real world.
One of the problems in small sample learning is few-shot image recognition, also known as few-shot image classification. This problem requires that the trained classification model can still accurately identify the category of the test sample even when there are very few training images for each category. Building a robust image recognition model using a small number of samples of the same category (even just one image) is a challenging problem.

\section{Experimental Dataset}
The dataset covers fine-grained classification, scene recognition, action recognition, and texture classification. 
The ImageNet~\cite{deng2009imagenet} dataset was constructed by Stanford University, The training set contains approximately 1.2 million images, while the validation set contains 50000 images. The Caltech101~\cite{fei2004learning} dataset covers common objects, animals, and scenes such as food, vehicles, furniture, etc. OxfordPets~\cite{parkhi2012cats} is a pet image dataset provided by the University of Oxford. StanfordCars~\cite{krause20133d} was released by Stanford University's Artificial Intelligence Laboratory in 2013. 
The Flowers102~\cite{nilsback2008automated} dataset was released by the University of Oxford in 2008, which includes 102 categories and a total of 8189 flower images. The Food101~\cite{bossard2014food} dataset is a commonly used fine-grained food classification dataset provided by ETH Zurich. The FGVCAircraft~\cite{maji2013fine} dataset is a fine-grained aircraft classification dataset that includes 100 aircraft categories.
The SUN397~\cite{xiao2010sun} dataset is a large-scale scene understanding dataset provided by Princeton University. The DTD~\cite{cimpoi2014describing} dataset is a texture classification dataset provided by the University of Oxford, containing 47 texture categories. The EuroSAT~\cite{helber2019eurosat} dataset is collected based on satellite images. The UCF101~\cite{soomro2012ucf101} dataset is a real-world action recognition dataset provided by the University of Florida. For the domain transfer dataset, we use four variants of ImageNet~\cite{recht2019imagenet,wang2019learning,hendrycks2021natural,hendrycks2021many}.

\begin{table}[h]
\footnotesize
\centering
  \begin{tabular}{lrrrr}
  \toprule  
  Dataset  & Classes  &  Train  &  Val  &   Test  \\
  \midrule  
 ImageNet~\cite{deng2009imagenet}  & 1,000 & 1.28 M &  N/A  & 50,000 \\
 Caltech101~\cite{fei2004learning}  & 100 & 4,128 & 1,649 & 2,465 \\
 OxfordPets~\cite{parkhi2012cats}  & 37 & 2,944 & 736 & 3,669 \\
 Flowers102~\cite{nilsback2008automated}  & 102 & 4,093 & 1,633 & 2,463 \\
 StanfordCars~\cite{krause20133d}  & 196 & 6,509 & 1,635 & 8,041 \\
 Food101~\cite{bossard2014food}  & 101 & 50,500 & 20,200 & 30,300 \\
 FGVCAircraft~\cite{maji2013fine}  & 100 & 3,334 & 3,333 & 3,333 \\
 SUN397~\cite{xiao2010sun}  & 397 & 15,880 & 3,970 & 19,850 \\
 EuroSAT~\cite{helber2019eurosat}  & 10 & 13,500 & 5,400 & 8,100 \\
 DTD~\cite{cimpoi2014describing}  & 47 & 2,820 & 1,128 & 1,692 \\
 UCF101~\cite{soomro2012ucf101}  & 101 & 7,639 & 1,898 & 3,783 \\
 
 ImageNet-V2~\cite{recht2019imagenet}  & 1,000 &  N/A  &  N/A  & 10,000 \\
 ImageNet-Sketch~\cite{wang2019learning}  & 1,000 &  N/A  &  N/A  & 50,889 \\
  ImageNet-A~\cite{hendrycks2021natural}  & 200 &  N/A  &  N/A  & 7,500 \\
  ImageNet-R~\cite{hendrycks2021many}  & 200 &  N/A  &  N/A  & 30,000 \\
\bottomrule
\end{tabular}

\caption{Training and testing datasets. This dataset consists of 11 image classification datasets and 4 variant datasets of ImageNet.}
\label{data}
\end{table} 

\section{Training}
These prompting methods have two important hyperparameters, namely the length of the prompt and the learning depth. In addition, in terms of the number of training iterations, there will be more training iterations for base to novel experiments, but fewer training iterations for cross-dataset generalization and domain generalization~\cite{fang2023extremely,fang2022out}, in order to avoid overfitting.
The underlying CLIP structure of these methods can be ViT-B/16, ViT-B/32, and ResNet. The experiments in this paper were conducted based on  ViT-B/16 for the purpose of fair comparison. 

\section{Generalizable Experiments Tasks}

\begin{itemize}
\item Generalization from base classes to novel classes. We divide a dataset into base classes and novel classes, and the model is trained only on the base class in the 16-shot setting, and evaluated on both the base class and novel class~\cite{zhou2022conditional}. We explore whether this architecture of fine-tuning prompts can have good performance in weakly generalized scenarios.

\item Generalization across datasets. To validate the effectiveness of models in cross-dataset migration tasks, we only trained the model on 16-shot ImageNet data, and then directly performed classification evaluations on 10 other unseen datasets. The purpose of this experiment is to explore whether models can successfully complete the recognition task of completely unseen data under strong generalization settings~\cite{zhou2022conditional}.

\item Domain generalization. We explore the robustness and generalization ability of models in weak generalization scenarios~\cite{fang2022semi,fang2021learning,fang2020open}. Similar to cross-dataset evaluation, we trained models on the ImageNet dataset with a 16-shot setting and directly evaluated its performance on four different variants of ImageNet (ImageNetA~\cite{hendrycks2021natural}, ImageNet-R~\cite{hendrycks2021many}, ImageNet Sketch~\cite{wang2019learning}, and ImageNetV2~\cite{recht2019imagenet}).

\end{itemize}

\section{Analysis}
Based on the current research status on few-shot prompt learning of CLIP for classification, the following issues and challenges can be summarized:

\begin{itemize}
    \item  In the experiment of generalization from base classes to novel classes, we found that the performance of base classes has almost reached a bottleneck, as shown in Table~\ref{design}, which may be due to the feature extraction ability of the visual branch. A large amount of work has been done using visual prompts to improve the performance of base classes, but the most advanced work currently indicates that there is not much room for performance improvement in this field~\cite{khattak2023self,zhang2024dept}.

    \item  In experiments using ImageNet as the source training data, the improvement in generalization performance was relatively small, especially in domain generalization experiments, as shown in Table~\ref{Domain}. The robustness of existing work needs to be improved in similar images.

    \item In most generalization experiments, prompting strategies are typically built around 16-shot settings. However, there are limited studies that explore small-sample fine-tuning scenarios like 1-shot, 2-shot, and 4-shot settings.

\end{itemize}

\section{Conclusion}
This article offers a concise summary of research on few-shot prompt learning based on CLIP, encompassing experimental findings and technical attributes of various methods. The primary aim of this review is to serve as a resource for newcomers in the research field focusing on generalizable prompting of CLIP across 15 datasets. Additionally, it aims to streamline the integration of this domain by researchers from diverse fields, providing insights into the methodologies and outcomes related to prompt learning with CLIP. This overview intends to aid researchers in understanding the advancements in leveraging CLIP for prompt learning. 
Further, we plan to investigate the potential of prompt learning in other tasks~\cite{tianfeng,yang2,lijinkai,yuzhou,liuzhe1,liuzhe2,liuzhe3,liuzhe6,liuzhe7} and scenarios~\cite{liwenxi,zhangwenyao,wangyuanze1,lihong,lihong2,lihong3,liushijia,liushijia2,liqing}.

{
    \small
    \bibliographystyle{ieeenat_fullname}
    \bibliography{arxiv}
}

\end{document}